# Toward Efficient Task Assignment and Motion Planning for Large Scale Underwater Missions

**Somaiyeh Mahmoud.Zadeh, David MW Powers, Karl Sammut, Amirmehdi Yazdani**

Centre for Maritime Engineering, Control and Imaging,
School of Computer Science, Engineering and Mathematics, Flinders University, Adelaide, SA 5042, Australia
somaiyeh.mahmoudzadeh@flinders.edu.au
david.powers@flinders.edu.au
karl.sammut@flinders.edu.au
amirmehdi.yazdani@flinders.edu.au

*Abstract-* An Autonomous Underwater Vehicle (AUV) needs to possess a certain degree of autonomy for any particular underwater mission to fulfil the mission objectives successfully and ensure its safety in all stages of the mission in a large scale operating field. In this paper, a novel combinatorial conflict-free-task assignment strategy consisting of an interactive engagement of a local path planner and an adaptive global route planner, is introduced. The method takes advantage of the heuristic search potency of the Particle Swarm Optimization (PSO) algorithm to address the discrete nature of routing-task assignment approach and the complexity of NP-hard path planning problem. The proposed hybrid method, is highly efficient as a consequence of its reactive guidance framework that guarantees successful completion of missions particularly in cluttered environments. To examine the performance of the method in a context of mission productivity, mission time management and vehicle safety, a series of simulation studies are undertaken. The results of simulations declare that the proposed method is reliable and robust, particularly in dealing with uncertainties, and it can significantly enhance the level of a vehicle's autonomy by relying on its reactive nature and capability of providing fast feasible solutions.

## 1. Introduction

Autonomous Underwater Vehicles (AUVs) are steadily becoming more widely used across a wide range of maritime applications. Recent breakthroughs in computer systems and sensor technologies have greatly expanded the range of missions that can be performed by AUVs. This widening operating range, however, is intertwined with the appearance of new complexities, for instance propagation of uncertainty and accumulation of data, transforming the underwater mission into a challenging scenario. Increasing the mission productivity and ensuring the vehicles safety are two main concepts that should be satisfied in all stages of the mission considering changes of environment. Hence, a vehicle must be able to autonomously perform effective and safe operation in such a severe environment. An efficient path planning strategy is a requisite to reach a satisfactory level of autonomy toward accomplishing underwater mission objectives. Efficient routing strategy promotes a vehicle's capability in task assignment, maneuver timing, and maneuver efficiency, while a real-time local path planner is capable of satisfying safe and optimal deployment in uncertain, dynamic, and cluttered ocean environments. Respectively, the prior investigations in this area can be considered along two distinct research fields; vehicle path planning, along with vehicle routing and task scheduling approaches.

The path planning strategies are known as efficient methods in vehicle guidance toward the destination encountering the local unexpected dynamic changes of the surroundings. Various strategies such as heuristic search algorithms [1-4], evolutionary and artificial intelligence based strategies [5-14] have been applied to the AUV motion planning problem in recent years. The Fast Marching (FM) algorithm is a heuristic approach utilized by Petres [3] to solve the path planning problem, which is efficient but computationally expensive compared to the A* algorithm. Afterward, Petres [4] employed an upgraded version of FM called heuristically guided FM (FM*) that keeps the accuracy of the FM along with the efficiency of the A*, but it is restricted to use linear anisotropic cost to attain computational efficiency. In general, the heuristic search algorithms are appropriate for real-time applications due to their high time complexity especially in large space problems. Evolution based optimization algorithms are a computationally faster approach that have been applied successfully to the path planning problem [5,6]. It has been proven that evolutionary algorithms are efficient methods in dealing with path planning as a Non-deterministic Polynomial-time (NP) hard problem and are capable of satisfying the time restrictions of real-time applications [5-8]. The Particle Swarm Optimization (PSO) [6,7], Genetic Algorithm (GA) [9-11], Differential Evolution (DE)[5], and Quantum-based PSO (QPSO) [12-13] are some commonly used optimization algorithms that have been successfully applied in the context of the unmanned vehicle path planning problem. Although most of these path planning techniques can accurately guide a vehicle toward the predefined destination, AUV-oriented applications still face many challenges especially across a large-scale operation area. In large scale operations, accurate estimation of the next state of the operating field (e.g., obstacles' behavior, etc.), far beyond a vehicle's current position and sensor coverage is computationally hard and experimentally impractical. On the other hand, in the case that the environment is populated with many waypoints and the vehicle is required to carry out a specific set of tasks (assigned to distance between waypoints in an operation network), a routing strategy would be more appropriate than a path planning strategy to prioritize tasks based on a global overview of the vehicles trajectory in the vast operating field.

Several research investigations have explored the problem of autonomous vehicle routing and task scheduling in a graph like operation environment. For instance, in the work by [14], a time-optimal conflict free route, relying on a few waypoints, was generated by a kind of an adaptive GA for operating in a large terrain. The heuristic nature of the PSO and GA algorithms has been investigated for the AUV routing and task allocation problem in the context of an NP-hard graph search problem [15]. The biogeography-based optimization (BBO) and PSO based route plan task assignment system is proposed in [16] for real-time AUV operation in a semi dynamic network. The merit of the proposed solutions given in [15,16] is that they are independent of the graph size which is not the case with most deterministic algorithms such as the proposed mixed integer linear programming (MILP) applied for governing multiple AUVs in [17]. The aforementioned strategies are mostly focused on task/target assignment and routing problems without considering the dynamicity of the environment and quality of deployment. The vehicles' routing strategies are dissimilar to path planning approach in terms of quality of deployment and are not able to handle rapid environmental changes.

This research contributes a reactive combinatorial conflict-free-task assignment scheme encompassing an adaptive global route and local path planning to cover the shortcomings associated with each of the aforementioned approaches. The route planner, in this context, is capable of generating a time efficient route with appropriate task assignment to ensure the AUV has a productive journey and ends the mission on-time, while the local path planner is responsible to generate an appropriate local path at a smaller scale to handle changes in the environment. In fact the local path planner and global route planner operate concurrently in parallel, while a constant interaction and information sharing is carried out between them. Their parallel execution accelerates the computation process and reduces unnecessary computation that would otherwise be caused by updates of whole terrain. A route re-planning capability is also incorporated to improve mission timing and improve the reactive ability of the AUV to environmental changes. The REMUS-100 vehicle is considered as the reference vehicle for this work. Referring to the kinematic and hydrodynamic model of the REMUS-100 and the corresponding constraints of the vehicle as given in [18], the problem formulation can then be developed. The constraints in the optimization problem reflect the natural constraints of the vehicle, and therefore, the solution obtained is realistic and applicable for real-word implementation.

The paper is structured as follows. In Section 2, the problem formulation is demonstrated. The PSO paradigm and its application for global route planning and local path planning is provided in Section 3. The simulation environment is presented in Section 4 and the results analyzed. In Section 5 the conclusions for this research are presented.

## 2. Problem Formulation of the AUV Routing and Local Path Planning Problem

The presented framework, in essence, is a decentralized model of a high-level route planner that searches for an appropriate set of tasks that can be accomplished in a limited time, and a low level local path planner, in which the model involves a decision maker core for determining the continuation of the path planning or need of the re-assignment of new tasks. This process ensures that the vehicle executes as many tasks as possible and provides a secure adaptive deployment toward the destination without unnecessary time expenditure. Therefore, the problem is formulated in two steps as follows.

### 2.1 AUV Routing Problem

Using a priori information about the underwater environment, first of all, the problem space should be transformed to a graph-like environment (depicted in Figure 2) in which all tasks are assigned to edges of the graph. The AUV starts its mission from one of the nodes called starting position $WP^s:(x_s,y_s,z_s)$ and tries to accomplish the maximum number of jobs (tasks along the graph edges) before reaching the predefined destination position $WP^D:(x_D,y_D,z_D)$ in a limited time. The positions of waypoints in the graph are initialized randomly using uniform distribution in 3D volume of $10000(x) \times 10000(y) \times 100(z)$ and connected by arbitrary edges $q_{ij}$. Each edge is assigned with a task and weighted by the priority of the corresponding task $Task_{ij}$: $w_{Tij}$. The goal is to find the best route through the given waypoints with maximum weight in a limited time that the

AUV's battery's capacity allows; this is a form of multi-objective optimization problem similar to the Knapsack-TSP. In the proceeding research, the AUV's route planning problem in 3D environment is formulated as follows:

$$\forall \begin{cases} WP^i : (x_i, y_i, z_i) \\ WP^j : (x_j, y_j, z_j) \end{cases} \rightarrow \exists q_{ij} : \begin{cases} d_{ij} = \sqrt{(x_j - x_i)^2 + (y_j - y_i)^2 + (z_j - z_i)^2} \\ t_{ij} = \dfrac{d_{ij}}{V_{AUV}} \\ w_{T_{ij}} \end{cases} \quad (1)$$

$$T_{Route} = \sum_{\substack{i=0 \\ j \neq i}}^{n} l q_{ij} t_{ij} = \sum_{\substack{i=0 \\ j \neq i}}^{n} l q_{ij} \left( \dfrac{d_{ij}}{V_{AUV}} \right), \quad l \in \{0,1\} \quad (2)$$

$$\max \left( \sum_{\substack{i=0 \\ j \neq i}}^{n} l q_{ij} w_{T_{ij}} \right), \quad l \in \{0,1\}$$

$$\min \left( \left| T_{Available} - T_{Route} \right| \right) \quad (3)$$

where $T_{Route}$ is the required time to pass the route, $T_{Available}$ is the total existing time, $l$ is the selection variable for any edge $q_{ij}$, $t_{ij}$ is the time to pass the distance $d_{ij}$ that includes weight $w_{Tij}$ of the corresponding task $Task_{ij}$. After a global route is generated, the path planner starts its operation of generating the optimum safe path among the existing waypoints in the global route.

### 2.2 AUV Local Path Planning Problem

In this section, the path planner aims to find the quickest collision-free path between waypoints. The dynamicity of the 3D operation field captures the motion of various static and freely suspended obstacles $\Theta$ with uncertain position and velocity, where the suspended obstacles are driven by current flow. An obstacle's velocity and uncertain position is modeled with a Gaussian distribution. Three different obstacles are modelled in this study to validate the efficiency of the path planner as follows:

i. **Static Object:** the position is initialized once in advance using normal distribution and does not change during the operation.

$$\forall \Theta_i \quad \exists \Theta_p^i, \Theta_r^i$$
$$\Theta_p^i \sim \mathbf{N}(0, \sigma^2), \quad \sigma^2 \approx \Theta_r^i \quad (4)$$
$$\Theta_p^i \in \left( WP_{x,y,z}^i, WP_{x,y,z}^{i+1} \right)$$

where, $\Theta_p$ is obstacle position that should be referenced to the location of start and target waypoints, $\Theta_r$ is its radius and initialized with a value in range of (0,100).

ii. **Static Uncertain Objects**: The radius of the position uncertainty for these objects changes with time within a specified boundary and its value in any time step is independent of its previous value.
$$\Theta_r^i \sim \mathbf{N}(\Theta_p^i, \sigma^2), \quad (5)$$

iii. **Moving Uncertain Obstacles:** these objects move along a direction motivated by current velocity, hence their radius and position changes iteratively according to (6):

$$\Theta_p(t) = \Theta_p(t-1) \pm U(\Theta_{p_0}, \sigma)$$
$$\Theta_r(t) = B_1 \Theta_r(t-1) + B_2 \times \mathbf{N}(0, \sigma) + B_3$$
$$B_1 = \begin{bmatrix} 1 & U_R^C(t) & 0 \\ 0 & 1 & 0 \\ 0 & 0 & 1 \end{bmatrix}, B_2 = \begin{bmatrix} 0 \\ 1 \\ 1 \end{bmatrix}, B_3 = \begin{bmatrix} 0 \\ 0 \\ U_R^C(t) \end{bmatrix} \quad (6)$$
$$U_R^C = |V_c| \sim \mathbf{N}(0, 0.3)$$

where $V_C$ is the current velocity, and $U_R^C$ is the defined uncertainty on objects motion.

#### 2.2.1 Path Construction Using B-Spline Method

To describe the corresponding B-Spline for local path generation, first the AUV model is demonstrated. An AUV has six degree of freedom (DoF) in its translational and rotational motion in Earth-fixed and Body-fixed coordinates [19], presented by Figure 1. The vehicle's earth-fixed state variables is presented in (7).

$$\eta : (X, Y, Z, \phi, \theta, \psi) \quad (7)$$

Where *X, Y, Z* are the AUV's position in the Earth-fixed coordinates; and $\phi$, $\theta$, $\psi$ are the Euler angles of roll, pitch, and yaw. The path planner in the proceeding research generates the potential trajectories $\wp_i=\{\wp_1,\wp_2,...\}$ using B-Spline curves shaped according to placement of number of control points $\vartheta=\{\vartheta_1, \vartheta_2,..., \vartheta_i,..., \vartheta_n\}$ in the specified operation window. Placement of these points play a substantial role in optimality of the path. Location of the control points $\vartheta^i:(x^i,y^i,z^i)$ in 3D space should be confined to predefined upper $U^i_\vartheta$ and lower $L^i_\vartheta$ bounds in the Cartesian coordinates as given bellow:

$$L_{\vartheta(x,y,z)} = [(x_0,y_0,z_0),(x_1,y_1,z_1),...,(x_{i-1},y_{i-1},z_{i-1}),...,(x_{n-1},y_{n-1},z_{n-1})]$$
$$U_{\vartheta(x,y,z)} = [(x_1,y_1,z_1),(x_2,y_2,z_2),...,(x_i,y_i,z_i),...,(x_n,y_n,z_n)]$$
$$x^i(t) = L^i_{\vartheta(x)} + Rand^x_i(U^i_{\vartheta(x)} - L^i_{\vartheta(x)})$$
$$y^i(t) = L^i_{\vartheta(y)} + Rand^y_i(U^i_{\vartheta(y)} - L^i_{\vartheta(y)}) \quad (8)$$
$$z^i(t) = L^i_{\vartheta(z)} + Rand^z_i(U^i_{\vartheta(z)} - L^i_{\vartheta(z)})$$

$$X(t) = \sum_{i=1}^{n} x^i B_{i,K}(t)$$
$$Y(t) = \sum_{i=1}^{n} y^i B_{i,K}(t) \quad (9)$$
$$Z(t) = \sum_{i=1}^{n} z^i B_{i,K}(t)$$
$$\wp^i_{x,y,z} = \sum_{x_s,y_s,z_s}^{|\wp|} \sqrt{(x^{i+1}-x^i)^2 + (y^{i+1}-y^i)^2 + (z^{i+1}-z^i)^2}$$

(9) gives the mathematical presentation of the B-Spline curve [9] generation in which a blending function of $B_{i,K}(t)$ and smoothness parameter of *K* are utilized to generate path curve coordinates *X, Y, Z* for the AUV position at time step *t* along the generated potential path $\wp$.

## 3. Particle Swarm Optimization and its Application to AUV Routing and Path Planning

The PSO is a popular optimization algorithm that offers a fast computation for solving different complex and multi objective problems. Its process starts with a population of particles in which each particle is characterized with position $\chi_i$ and velocity $v_i$ that is updated iteratively. Particles correspond to candidate solutions in the search space and should be evaluated iteratively according to a predefined cost functions. Particles memorize their previous state value $\chi(t-1)$, their experienced best position of $\chi^{P-best}$ and also the global best $\chi^{G-best}$ positions in the population. More detail can be found in [20]. Particle velocity and position are updated as follows:

$$v_i(t) = \omega v_i(t-1) + c_1 r_1 [\chi^{P-best}_i(t-1) - \chi_i(t-1)] + c_2 r_2 [\chi^{G-best}_i(t-1) - \chi_i(t-1)]$$
$$\chi_i(t) = \chi_i(t-1) + v_i(t) \quad (10)$$

where $c_1,c_2$ and $r_1,r_2$ are acceleration coefficients and independent random numbers, respectively. The $\omega$ is the inertia weight and balances the local and global search. The PSO is here used for both AUV routing and path planning purpose as it is a strong algorithm for handling complex scaling and multi-objective problems. As this algorithm was developed to operate in a continuous space, it is appropriate for solving the path planning problem that is a continuous natured problem. Furthermore, to adapt this algorithm for solving discrete problems such as knapsack or vehicle routing problem this research proposes a priority based feasible route generation approach on the underlying search space (provided in Section 3.2). These modifications increase the speed of the algorithm in finding optimum solution and prevent trapping in a local optima.

### 3.1 *PSO and Path Planning Approach*

The path planning aims to generate the fastest collision-free path between two points. This research takes the advantages of the B-Spline method and the PSO algorithm to generate desirable path curves, where each particle in the swarm is assigned by coordinates of a potential path. The position and velocity of the particles are assigned with the coordinates of the control points $\vartheta$ of the B-spline. As the PSO iterates, particles move toward their respective local attractor according to their individual and swarm search. The process of PSO based path generation is provided by Figure 2. After candidate paths (solutions) are generated their cost/fitness should be evaluated, where in this study path cost function is defined proportional to the path time $T_{path\text{-}flight}$ as follows:

$$\forall \wp, \quad \wp \approx \sum_{1}^{|\wp|} \vartheta_{j+1} - \vartheta_j,$$
$$\wp^j_{x,y,z} = \sum_{x_s,y_s,z_s}^{|\wp|} \sqrt{(\vartheta_{x(j+1)} - \vartheta_{x(j)})^2 + (\vartheta_{y(j+1)} - \vartheta_{y(j)})^2 + (\vartheta_{z(j+1)} - \vartheta_{z(j)})^2} \quad (11)$$

$$T_{path-flight} = \sum_1^n t_i = \sum_1^{|\wp|} \Gamma_{3D}\left\{\vartheta_{t_i}^i\right\} = \sum_1^{|\wp|} \frac{\left|\vartheta_{i+1}^{t_{i+1}} - \vartheta_i^{t_i}\right|}{|V_{AUV}|} \qquad (12)$$

$$Cost_\wp \approx \min(T_{path-flight})$$
s.t.
$$\forall j \in \{0,...,|\wp|\} \Rightarrow \vartheta_j^{t_i} \notin \Theta(t_j) \cup \Gamma_{3D} \quad and \quad j \notin \bigcup_{N\Theta} \Theta(\Theta_p, \Theta_r) \qquad (13)$$

where, $Cost_\wp$ is the path cost function. It is assumed the vehicle moves with constant linear velocity $V_{AUV}$. The generated $\wp$ is penalised for colliding with any obstacle $\Theta_{(N\Theta)}$, where $N_\Theta$ is the number of obstacles. This process in general motivates particles toward the feasible solution space.

### 3.2 PSO and AUV Routing

Regarding the combinatorial nature AUV's routing problem which is combination of TSP and Knapsack problems, there should be a compromise between remaining time, maximizing number of higher weighted tasks, and guaranteeing on-time completion of the mission. The AUV route planner simultaneously tends to determine the appropriate route in graph-like terrain while trading-off between maximizing the total collected weights along the route and satisfying available time threshold. The solution by the route planner generally contains optimum set of jobs/waypoints, in which the tasks are prioritized according to their weight. A general overview of the PSO mechanism on AUV routing is provided by Figure 3. Suitable coding of the particles by route solutions is the most critical factor affecting the algorithm's efficiency and optimality of the generated solutions.

```
Initialize each particle by random velocity and position in following steps:
    – Assign B-spline control points ϑi as particle position χi
    – Initialize each particle with random velocity υi in range of predefined bounds Uiϑ, Liϑ
    – Choose appropriate parameters for the population size (imax)
    – Set the number of control-points (m) that used to generate the B-Spline path
    – Set the maximum number of generations (tmax)
    – Initialize acceleration coefficients
    – Initialize each particles current best position χiP-best(1), at first iteration t=0
    – Set the χiP-best(1) as the best fitted particle χiG-best(1) so far, at iteration t
    – evaluate the initial population according to given cost function
BEGIN
For t=1 to tmax
    For i=1 to imax
        update particle velocity using eq(10)
        update particle position using eq(10)
        evaluate particles:
        χiP-best(t) = { χiP-best(t-1),  if {Costφ(χi(t)) ≥ Costφ(χiP-best(t-1))}
                     { χi(t),          if {Costφ(χi(t)) < Costφ(χiP-best(t-1))}
        χG-best(t) = argmin      Costφ(χiP-Best(t))
                    1≤i≤χP-Best
        Updated the particles personal best χiP-best
        Updated the swarm's global best position χG-best at iteration t
    End
    evaluate the cost of each candidate particle (path) according to given cost function
End
Output χG-best as the optimal solution (path)
END
```

Figure 2. PSO optimal path planning pseudo code

#### 3.2.1 Particle Encoding (Initialization Phase of PSO-Route Generation)

A particle in the proposed method corresponds to a feasible route vector including a sequence of jobs/waypoints. The solution vectors takes a variable length with maximum size equal to the total number of existing waypoints in the graph. A feasible route must meet the following criteria:

[1] It should be started and ended with the index of the predefined start and target waypoints.
[2] It should not include connections (edges) that are not given in the graph.
[3] It should not visit the same waypoint multiple times.
[4] It should not visit an edge more than once.
[5] It should not take longer than the maximum available time.

A priority-adjacency based method has been applied in this study to produce feasible initial solutions/routes, in which guiding information is added to each waypoint of the graph at the initial phase. Waypoints are selected according to their corresponding adjacency relations in the graph and their priority value (the adjacency matrix represents relations and edges in a graph) and to keep the solutions feasible some modifications have been done as follows:

- The priority vector is initialized with random uniform value ~**U**(-100,100) and each waypoint in the graph takes a priority value from priority vector.
- Visited waypoints in a route are assigned with a large negative priority value to ignore visited nodes.
- The included edges in the route are excluded from the adjacency matrix.
- Adjacency connections are used for adding waypoints to route, so that non-existing edges will be excluded in the route.

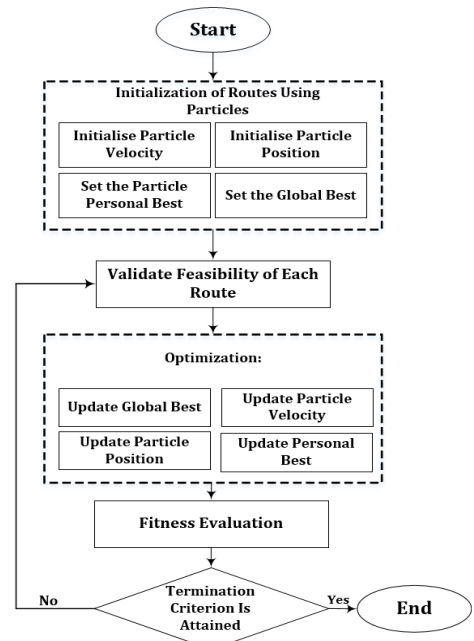

Figure 3. The process of PSO on AUV route planning

An example of the proposed feasible route generation process is proposed by Figure 4 using a sample graph with 18 waypoints and its corresponding adjacency matrix ($Ad$) and priority array ($U_i$). In this process, the first waypoint is added to the route as the start position. Then the priority of the connected nodes to the first waypoint is considered (using $Ad$ matrix), which in the example sample graph is the sequence of {2,3,4,5}, and the highest priority node is added to the sequence. This procedure is repeated until visiting the destination node (node 18 in this graph).

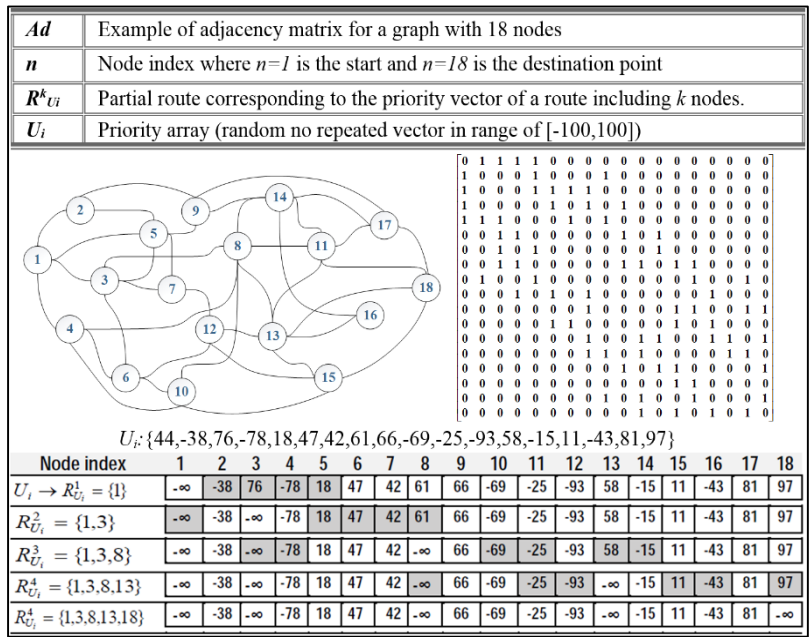

Figure 4. Process of feasible route generation by topological information of a sample

Termination of the PSO operation is defined according to completion of the iterations or when fitness of the generations get reduced iteratively. The important step in fitness evaluation is defining an efficient cost function, so that the algorithm tends to compute the cost value for each route to find the best fitted route with the minimum cost (given by Section 3.2.3).

### 3.2.2 Adaptive Route Re-planning

The local path planner operates inside the global route planner and concurrently tends to safely guide the vehicle through the waypoints. If the local path time ($T_{path\text{-}flight}$) takes longer than the expected time for traversing the edge $q_{ij}$ ($T_{Expected}$) due to encountering unexpected changes in the environment, the initial route loses its optimality because of the wasted time and battery consumption; so the $T_{Avaliable}$ is updated and route re-planning is carried out to compute a new optimum route according to mission updates. In the re-planning process the visited edges are omitted from the graph and the search space shrinks. Then the current node is considered as the new starting waypoint for both planners.

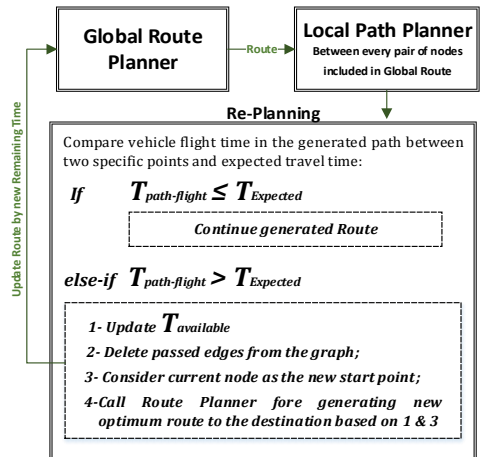

Figure 5. Proposed route re-planning process

Using such an adaptive route re-planning compensates the time dissipation and improves the reactive ability of the vehicle considering the environmental changes. The re-planning process is presented in Figure 5. To clarify this combinational process an example is provided in Figure 6. Considering the initial optimum route as the sequence {S-1-7-3-4-5-6-9-10-13-17-19-D} in Figure 6, the path planner tends to generate a trajectory to safely guide the AUV through the waypoints. The re-planning condition (given by Figure 5) is investigated after each process of local path planner and in the case that re-planning is required, the mentioned updates presented in Figure 5 are applied and the route planner is recalled to find a new optimum route considering mission updates (e.g. new optimum route:{9-8-10-16-D}). This process repeats until vehicle reaches the required waypoint.

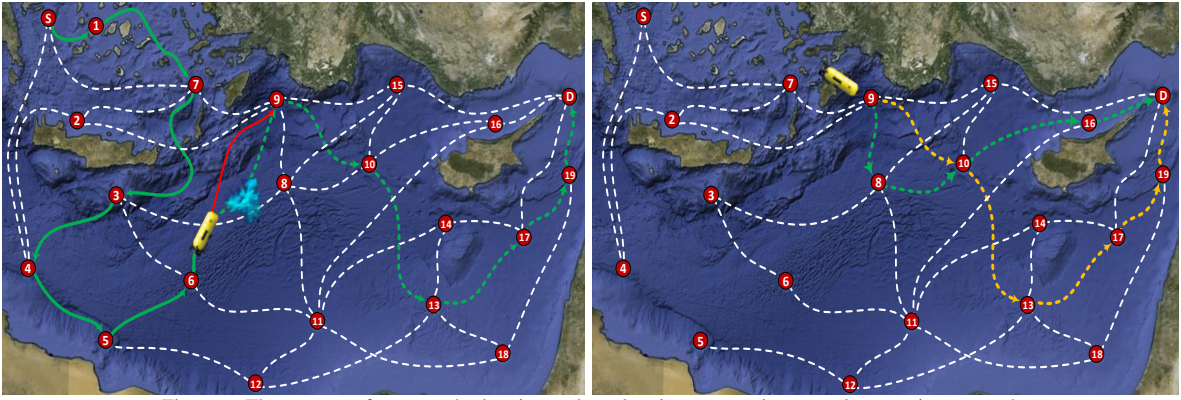

Figure 6. The process of route-path planning and re-planning process in a sample operation network

### 3.2.3 Global Route Optimality Evaluation

The route cost is directly influenced by the cost of the generated local path $Cost_\wp$. The route planner searches for an appropriate solution in the sense of the best composition of path, route, and task cost defined as follows:

$$Cost_{Route} = \left| T_{Available} - \sum_{\substack{i=0 \\ j \neq i}}^{n} lq_{ij}\left(Cost_{\wp ij}\right) \right| + \sum_{\substack{i=0 \\ j \neq i}}^{n} lq_{ij} \times w_{T_{ij}} \quad , \quad l \in \{0,1\} \tag{14}$$

where, $Cost_{Route}$ is the route cost that is defined as a function of $Cost_\wp$.

## 4. Simulation

The main purpose of the simulation experiments in this paper, is evaluating the performance of entire combinational model in terms of increasing mission productivity (task assignment and time management) and guaranteeing safe maneuver during the mission. To validate the proposed strategy, first the efficiency of the local path planner is individually assessed; afterward, its performance in context of the global route planner is investigated.

### 4.1 Simulation Results for PSO Based Local Path Planner

During transition between two points, the local path plan effectiveness can be compromised by dynamic changes of the environment. The purpose of path planner is to minimize the path time, and to avoid collision. It is assumed the vehicles is moving with constant speed of (3 *m/s*). The operating environment is modeled as a realistic underwater volume covered by several uncertain static and suspended obstacles. Three different scenarios are simulated for evaluating the efficiency of the local path planner. In the first one, the AUV starts its operation in a static environment and it is required to pass the shortest collision free distance. In the second scenario, the robustness of the method is tested in an environment covered by uncertain static obstacles. Finally, the terrain becomes more complicated by adding the floating objects into account. The complexity of the terrain is increased gradually to test the path planner's performance in all probable conditions. The PSO is configured to run 100 iterations and 80 particles, with expansion-contraction coefficients of 1.5 and 2. The number of control points is set on 6. The performance of the algorithm for all scenarios proposed is shown in Figures 7 to 9 for different number of obstacles, where each obstacle is generated randomly according to the equations presented in Section 2.2.

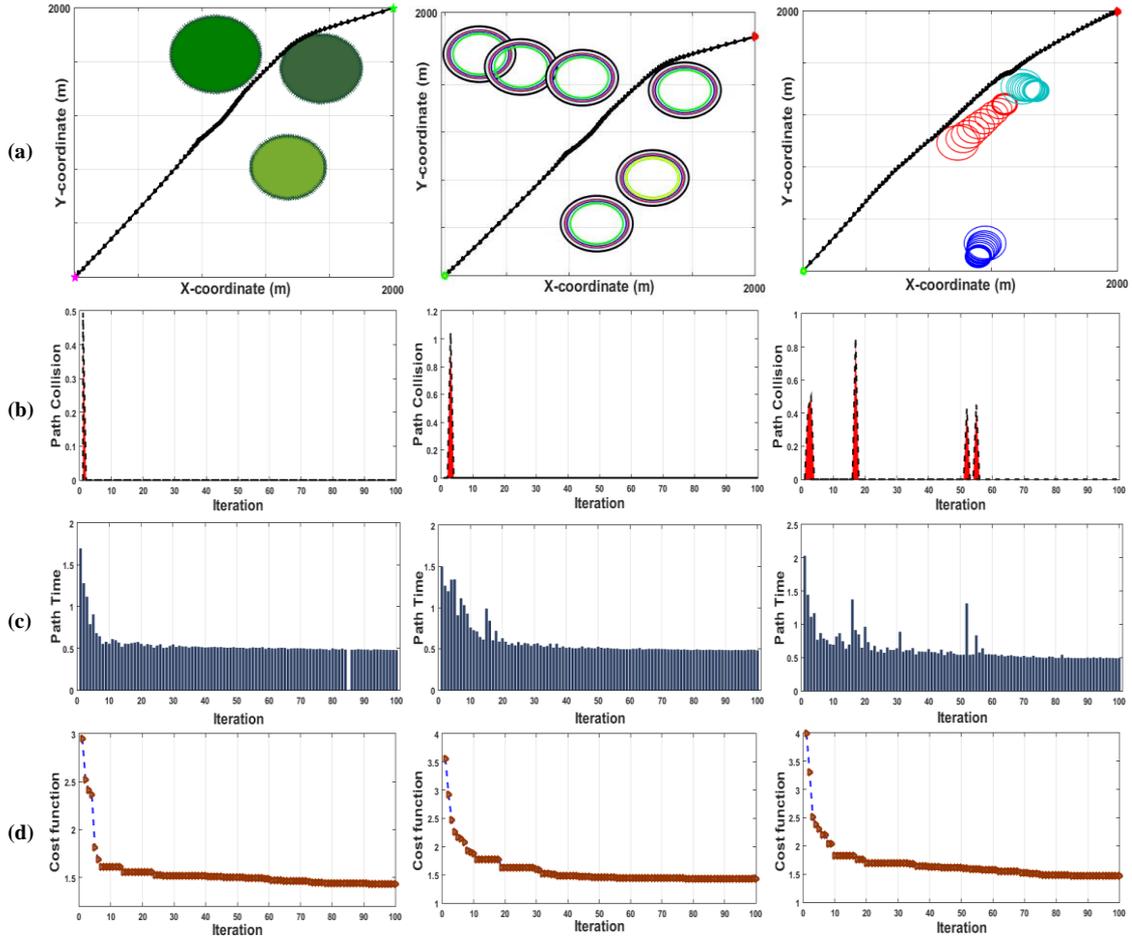

Figure 7. (a) The path planning performance in avoiding **known static obstacles** in the 1st scenario. (b) Iterative variation of the collision violation. (c) Iterative variation of path time. (d) Iterative variation of path cost.

Figure 8. (a) The path planning performance in avoiding *uncertain static objects* in the 2nd scenario. (b) Iterative variation of the collision violation. (c) Iterative variation of path time. (d) Iterative variation of path cost.

Figure 9. (a) The path planning performance in avoiding *uncertain suspended objects* in the 3rd scenario. (b) Iterative variation of the collision violation. (c) Iterative variation of path time. (d) Iterative variation of path cost.

The gradual growth of collision edges of each uncertain objects is shown in *Figure* 8(a) and 9(a) in which the uncertainty growth is assumed linear with each iteration. As presented in *Figures* 7(b), 8(b), and 9(b) the algorithm efficiently manages the trajectory to eliminate the collision penalty within 100 iterations. It is evident from *Figure* 7(a), 8(a), and 9(a) that the algorithm accurately satisfies the collision constraints encountering all types of obstacles. The path travel time is an important performance index that is proportional to path cost and should be minimized. It is inferable from *Figure* 7(c,d), 8(c,d), and 9(c,d), the algorithm accurately tends to minimize path travel time and path cost over 100 iterations, while the performance of the algorithm is almost stable against the increasing complexity of the environment.

### 4.2 *Simulation Results for Reactive Global Route Planner*

The main goal is to make maximum use of the mission time, to increase the mission performance by optimum routing, to ensure on-time ending of the mission; and concurrently providing safe motion to the final destination. The route, as a sequence of arranged tasks, provides an overview of the operation area that an AUV should fly through. On the other hand, dynamic changes of the environment is covered by the path planner, which operates concurrently in an inner layer. The local path planner operates as an inner layer of the global route planner and the output of each planner concurrently feeds to the other one. The entire combinational scheme should be fast enough to cope with dynamic changes, handle environmental updates, and carry out prompt re-planning to compute new order of tasks according to new updates. To this end, some performance factors of total violation, cost, CPU time, mission time, and remained time, are investigated through six individual experiments to evaluate the performance and stability of the model in providing safe motion and beneficent task assignment. The operation graph is configured with 30 nodes and 985 edges involving a fixed set of 10 tasks specified with their priority and randomly distributed in $10 \times 10\ km^2$ (*x-y*), 100 *m* (*z*) space. The total available time is set by $T_{Available}=9000\ sec=2.5\ hours$. The AUV starts its operation at point $WP^1$ and end its operation at $WP^{30}$. The terrain is modelled to be randomly covered by various uncertain objects. The PSO configured with the same initial conditions of the local path planner. This simulation is implemented on a desktop PC with an Intel i7 3.40 GHz quad-core processor using MATLAB® R2014a. The full process of one experiment out of the six separate missions (given by Figure.10 to 12) is explained in Table.1(A-B) to clarify the operation flow of the proposed combinational strategy across all stages of a particular mission toward addressing the mentioned objectives.

Table 1. An overview of the process of the combinational model in one mission scenario

| | | Call NO | $WP^S$ | $WP^D$ | Task NO | Route Weight | Route Cost | CPU Time | $T_{Available}$ | $T_{Route}$ | Valid | Route Sequence |
|---|---|---|---|---|---|---|---|---|---|---|---|---|
| A | Global Route Planner | 1 | 1 | 30 | 7 | 18 | 2.2097 | 76.15 | 9000 | 8597 | Yes | 1-23-16-28-3-15-17-30 |
| | | 2 | 28 | 30 | 7 | 21 | 1.9647 | 68.61 | 6472.2 | 6118 | Yes | 28-18-9-3-24-7-29-30 |
| | | 3 | 18 | 30 | 5 | 23 | 2.3386 | 70.81 | 5952.3 | 5612 | Yes | 18-5-25-21-8-30 |
| | | 4 | 8 | 30 | 3 | 20 | 1.8659 | 67.16 | 1871.6 | 1693 | Yes | 8-20-7-30 |
| | | 5 | 20 | 30 | 1 | 7 | 1.4903 | 59.07 | 884.5 | 819 | Yes | 20-30 |

| | | Route ID | PP Call | Edges | Violation | Path Cost | CPU Time | $T_{path}$ | $T_{Expected}$ | $T_{Available}$ | Replan Flag | PP Flag |
|---|---|---|---|---|---|---|---|---|---|---|---|---|
| B | Local Path Planner | Route-1 | 1 | 1-23 | 0.00000 | 0.328 | 36.31 | 1476 | 1514.3 | 7524 | 0 | 1 |
| | | | 2 | 23-16 | 0.00028 | 0.287 | 31.25 | 565 | 683 | 6959 | 0 | 1 |
| | | | 3 | 16-28 | 0.00000 | 0.743 | 44.38 | 486.7 | 333.8 | 6472.2 | 1 | 0 |
| | | Route-2 | 1 | 28-18 | 0.00000 | 0.109 | 39.6 | 519.6 | 376.3 | 5952.3 | 1 | 0 |
| | | Route-3 | 1 | 18-5 | 0.00000 | 0.120 | 39.60 | 1406 | 1492.3 | 4546.1 | 0 | 1 |
| | | | 2 | 5-25 | 0.00000 | 0.101 | 44.49 | 394.6 | 486.3 | 4151.5 | 0 | 1 |
| | | | 3 | 25-21 | 0.00000 | 0.360 | 41.45 | 1192.6 | 1345.3 | 2959.3 | 0 | 1 |
| | | | 4 | 21-8 | 0.00000 | 0.416 | 40.56 | 1087.3 | 900.3 | 1871.6 | 1 | 0 |
| | | Route-4 | 1 | 8-20 | 0.00000 | 0.319 | 53.93 | 986.8 | 847.8 | 884.5 | 1 | 0 |
| | | Route-5 | 1 | 20-30 | 0.00000 | 0.233 | 42.19 | 796.4 | 818.6 | **87.8** | 0 | 0 |

The mission commences with operation of the global route planner to produces a valid optimum route in order to take maximum use of available time ($T_{Route} \leq T_{Available}$). Considering Table.1(A), the initial global route includes 7 tasks, weight of 18 and cost of 2.2097 with mission time of $T_{Route}=8597$ (s); then, the local path planner generated shortest and safest path between the set of nodes listed in the initial route. Considering Table.1(B), the local path planner generated a trajectory between the first pair of nodes (1-23) with total cost of the 0.328, and time of $T_{path}=1476$, which is smaller than expected travel time $T_{Expected=}1514.3$. When $T_{path}$ is smaller than $T_{Expected}$ the re-planning flag is zero which means that the initial route is still valid and optimum, so the vehicle is allowed to follow the next pair of waypoints included in initial optimum route. The $T_{Available}$ gets updated after each operation of the local path planner. The second pair of waypoints (23-16) is shifted to the local path planner and the same process is repeated. However, path planning in (16-28) took longer than $T_{Expected}$, so the re-planning flag turned on, the $T_{Available}$ is updated and visited edges (1-12, and 23-16) are deleted from the graph. Afterward, the global route planner is recalled to produce a new optimum route from the current waypoint $WP^{28}$ to the $WP^{30}$ according to new updates. In the results shown in Table.1, the local path planner is called 10 times within 5 optimum routes. This synchronization among the global and local motion planners continues until vehicle reaches its destination (success) or $T_{Available}$ gets a minus value (failure: vehicle runs out of battery). The final route passed by the vehicle in this mission through the 5 route re-planning and 10 path planning is the sequence of waypoints including {1-23-16-28-18-5-25-21-8-20-30} with total cost of 1.878, total weight of 22, and total time of 8912. From simulation results in Table.1(A), it is noted that in all cases route travelling time obtained by the PSO is smaller than total available time and the violation value for all solutions is equal to zero thus confirming feasibility of the produced route and that the model respects the defined constraints.

The best result for the model is completion of the mission with the minimum remained time ($T_{Remained}$) that means maximizing the use of available time. The results provided in Table.1(A) confirm that the model acts efficiently in terms of maximizing mission productivity by making maximum use of the available time as $T_{Route}$ approaches $T_{Available}$. Referring to Table.1(B), the remaining time is 87.8 second out of the total mission time available $T_{Available}= 2.5$ hr, which is considerably approached to zero. Considering that reaching the destination, is the biggest concern for a vehicle's safety, it is more important than maximizing the vehicle's productivity, a big penalty value is assigned to the global route planner to strictly prevent generating routes with $T_{Route}$ bigger than $T_{Available}$.

In this research, the computational time is another important performance factor that should be investigated for the path-route planner because they operate concurrently, thus a large computational time for either of them is detrimental to the routine flow of the whole system. The robustness of the model in satisfying the addressed objectives is evaluated by testing six missions through six individual simulations all with equivalent initial conditions based on realistic underwater mission scenarios. These results are presented in Figure 10, 11 and12.

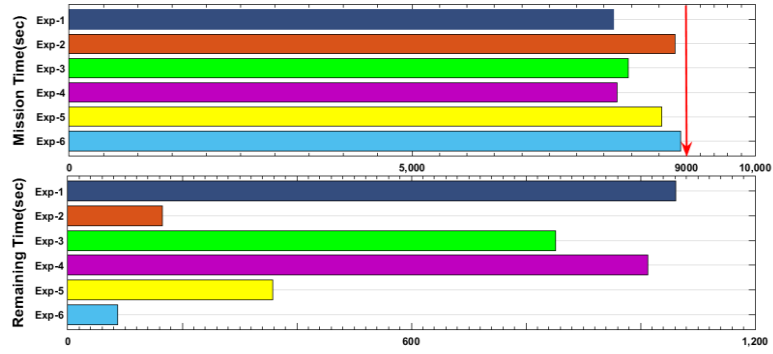

Figure 10. Performance of the combinational model in mission time management

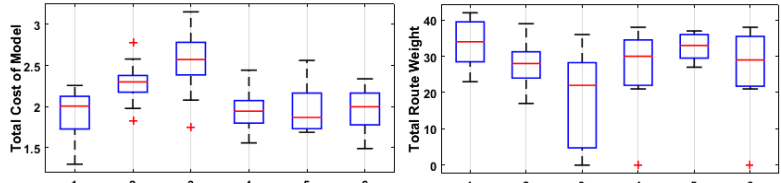

Figure 11. Stability of the global route planner's cost variation of and variation of total obtained weights in 6 individual experiments

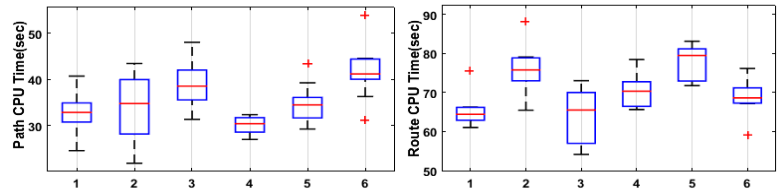

Figure 12. Stability of the Global route and local path planner in computational time variation

From the simulation results given in Figure 10 it can be seen that the proposed architecture is capable of making maximum use of available mission time as apparently the mission time in all experiments approaches the $T_{Available}$ and meet the above constraints denoted by the upper bound of 2.5 *hours*, which presents the stability of the model in mission time management. It is evident from Figure 11 that the average cost variation of the global route planner in each experiment is centralized over the total cost of the final route produced in the corresponding mission. It is also noteworthy to mention from analyzing the results in Table.1(A-B) that the proposed methodology takes a very short computational time which makes it highly suitable for real-time application. The stability of the variability of computational time for both global and local planners as shown in Figure 12, also confirms the applicability of the model for real-time implementation. More importantly, the violation percentage for both planners presented in Table.1, reveals the robustness of the model to the variations of the operation network parameters and environmental conditions, so that the vehicle can meet the dynamic challenges of the environment and guarantee safe deployment and on-time completion of the mission.

## 5. Conclusion

This paper focuses on developing a reactive combinational conflict-free-task assignment method based on reactive global route planning and collision-free local path planning that ensures the consistent situational awareness of the environment and guarantees a real-time implementation. Exploiting a priori knowledge of the underwater environment, the global route planner generates a time efficient route with best sequence of tasks to be carried out to ensure the AUV has a successful operation and reaches its destination on-time as it is an obligatory requirement for vehicles safety. Along with an efficient route planner, a local path planner is produced in smaller scale to safely guide the vehicle through the specified waypoints in the global route with minimum time/energy cost. The employed PSO-based local path planner is efficient and fast enough to rapidly react to changes of the environment and generate a collision-free optimum path. Route re-planning, meanwhile, is performed to reduce wasted time in dealing with unexpected changes of the environment and consequently promote robustness and reactive ability of the AUV. On the other hand, stability of the hybrid planner in time management is the most critical factor representing robustness of the method. The value of the remaining time upon completing the mission should be minimal but should not be equal to zero. Minimizing the remaining time maximizes the mission productivity. The results obtained from analyzing the six different missions, demonstrates the inherent robustness and effectiveness of the proposed scheme in enhancing a vehicle's mission productivity and mission time management.